\documentclass[review]{elsarticle}

\usepackage{lineno,hyperref}

\usepackage{times}
\usepackage{epsfig}
\usepackage{graphicx}
\usepackage{amsmath}
\usepackage{amssymb}
\usepackage{multirow}
\usepackage{array}
\usepackage{booktabs}
\usepackage{color}

\usepackage{amsmath}
\usepackage{amssymb}

\newcommand{\myfigref}[1]{Figure \ref{#1}}
\newcommand{\mytabref}[1]{Table \ref{#1}}
\usepackage{threeparttable}
\usepackage{bm}
\usepackage{subfigure} 
\usepackage{graphicx}
\usepackage{booktabs}
\usepackage[dvipsnames]{xcolor} 
\usepackage{float}

\begin{document}

\begin{frontmatter}

\title{Towards Reliable Domain Generalization: \\ A New Dataset and Evaluations}


\author[mymainaddress,mysecondaryaddress]{Jiao Zhang\corref{mycorrespondingauthor}}
\ead{zhangjiao2019@ia.ac.cn}
\author[mymainaddress,mysecondaryaddress]{Xu-Yao Zhang}
\ead{xyz@nlpr.ia.ac.cn}
\author[mymainaddress,mysecondaryaddress]{Cheng-Lin Liu}
\ead{liucl@nlpr.ia.ac.cn}

\address[mymainaddress]{State Key Laboratory of Multimodal Artificial Intelligence Systems (MAIS), Institute of Automation of Chinese Academy of Sciences, Beijing 100190, China}
\address[mysecondaryaddress]{School of Artificial Intelligence, University of Chinese Academy of Sciences, Beijing 100049, China}



%

\begin{abstract}
There are ubiquitous distribution shifts in the real world. However, deep neural networks (DNNs) are easily biased towards the training set, which causes severe performance degradation when they receive out-of-distribution data. Many methods are studied to train models that generalize under various distribution shifts in the literature of domain generalization (DG). However, the recent DomainBed and WILDS benchmarks challenged the effectiveness of these methods. Aiming at the problems in the existing research, we propose a new domain generalization task for handwritten Chinese character recognition (HCCR) to enrich the application scenarios of DG method research. We evaluate eighteen DG methods on the proposed PaHCC (Printed and Handwritten Chinese Characters) dataset and show that the performance of existing methods on this dataset is still unsatisfactory. Besides, under a designed dynamic DG setting, we reveal more properties of DG methods and argue that only the leave-one-domain-out protocol is unreliable. We advocate that researchers in the DG community refer to dynamic performance of methods for more comprehensive and reliable evaluation. Our dataset and evaluations bring new perspectives to the community for more substantial progress. We will make our dataset public with the article published to facilitate the study of domain generalization.
\end{abstract}


\end{frontmatter}


\section{Introduction}
Deep neural networks have achieved remarkable performance on many classical datasets, but they may suffer from serious performance degradation in real-world applications. One reason is the data distribution shift. Specifically, in the open world, the distribution of test data is usually inconsistent with that of training data due to unpredictable environmental changes and the bias introduced in the data collection process \cite{torralba_unbiased_2011}. Tackling distribution shifts among training and test data, which is known as domain generalization (DG) or out-of-distribution generalization (OODG), is still an open problem.

In the last decade, researchers in the DG community proposed a plethora of methods to enhance the model's generalization ability from different perspectives \cite{ZhouLQXL23,wang2022generalizing}. However, there is still no clear conclusion on what the most effective solution is. In particular, recent empirical benchmarks \cite{GulrajaniL21,KohSMXZBHYPGLDS21} challenged the progress of domain generalization methods. The authors of the work \cite{GulrajaniL21} tested fourteen methods on seven datasets under the same experimental conditions. It shows that none of the tested methods simultaneously surpass the baseline model (ERM).The work \cite{KohSMXZBHYPGLDS21} points out that current datasets widely used in the community under-represent distribution shifts in real-world deployments. To address this gap, they present a curated benchmark of 10 datasets that reflect diverse distribution shifts in real-world applications. However, the Chinese character recognition scenario has not been touched.

To facilitate the practicability and versatility of DG research, here we construct a new domain generalization task for the Chinese character recognition scenario. Compared with English characters, Chinese characters are more difficult to recognize because of their large categories and complex structures. For the handwritten Chinese character recognition task, collecting handwritten data requires vast manual and material resources. Besides, due to the inconsistency of writing styles between writers, the generalization between different writers is a problem usually considered in the previous works \cite{zhang2011style,zhang2012writer}. Considering the rapid development of domain generalization in recent years, here we propose a generalization task from synthetic printed Chinese characters to scanned handwritten ones, which uses only synthetic printed data to guide the model to learn to deal with different fonts, and hope that the trained model can directly generalize to the recognition of scanned handwritten data. For this task, we construct a Non-I.I.D.\footnote{Non-I.I.D.: Non-Independent and Identically Distributed} image dataset called PaHCC (Printed and Handwritten Chinese Characters), which contains 280 printed fonts and 720 handwritten fonts. We divide the synthetic printed data into three domains according to the font types and regard all scanned handwritten data as one domain.

On the proposed PaHCC dataset and the DomainNet dataset, we conduct extensive evaluation experiments. By testing multiple methods contained in DomainBed \cite{GulrajaniL21}, we show that Chinese character recognition differs from object recognition on distribution shifts and existing methods cannot handle this task well. In addition, we evaluated 18 methods on both datasets under a series of ``dynamic" DG settings. Our experiments reveal more properties of DG methods and indicate that the common leave-one-domain-out protocol is unreliable. Further, we propose two ``dynamic" criteria, namely the ``Domain +" criterion ( performance variation when increasing the number of training domains) and the ``ST (swap test)" criterion (performance variation when selecting different training domains without changing the number of training domains), which can assist the original criteria for more comprehensive and reliable evaluations.

In summary, our contributions are as follows:
\begin{itemize}
  \item We propose a practical and novel domain generalization task for handwritten Chinese character recognition, which enriches the application scenarios of domain generalization research.
  \item We construct a large-scale Chinese character generalization dataset called PaHCC (Printed and Handwritten Chinese Characters), containing 280 printed fonts and 720 handwritten fonts, for the study of printed-to-handwritten Chinese character recognition.
  \item We evaluate multiple DG methods contained in DomainBed \cite{GulrajaniL21} on the proposed PaHCC dataset. Our experiments show that existing DG methods cannot handle this task well, and there is still much room for improvement.
  \item By a series of dynamic DG tests, we show that the common leave-one-domain-out protocol is unreliable and propose two new criteria for more comprehensive and reliable evaluations.
\end{itemize}

\section{Domain Generalization}
\label{DG_introduction}
In this section, we classify and review the existing methods in the DG community. For a comprehensive understanding, we also briefly introduce commonly used public datasets and the evaluation criteria in the study of domain generalization. On our experiments, we evaluate and analyze the selected eighteen algorithms.

\subsection{Methods}
In recent years, researchers have proposed a plethora of domain generalization methods from different perspectives, which can be roughly grouped into three categories: \emph{data manipulation}, \emph{representation learning}, and \emph{learning strategies}.

In over-parameterized deep neural networks (DNNs), data augmentation is often used as a data-level regularization to alleviate the over-fitting of the model and improve generalization \cite{Ian_deep_learning2016}. Due to its significant impact on the performance of data-driven DNNs, many works focus on enriching the diversity of training data by various \emph{data manipulation} to benefit DG. Earlier work perturbs the input samples using the gradient of the classification loss \cite{ShankarPCCJS18,Volpi_generalizing_2018} or synthesize new training samples with the help of generative models \cite{zhou_deep_2020,RobeyPH21}. Recently, the feature augmentation \cite{LiLLGFH21,zhou_domain_2021,KangLKK22} of raw data in the feature space has attracted more and more attention due to the simple operation and small computation cost. In addition, some methods \cite{0001GXL21, XuZ0W021} enhance the data by decoupling and mixing the content and context information of different samples in the frequency domain and then transforming it back to the spatial domain.

\emph{Representation learning} works cope with distribution shift mainly by learning domain-agnostic representation. Domain alignment is a common approach that aligns the marginal distributions of source domains \cite{Muandet_domain_2013,LiPWK18} or class-conditional distributions across source domains \cite{LiTGLLZT18,li_conditional_2018} and hopes that the learned invariance between source domains can extend to unseen test domains. Learning disentangled representation \cite{peng_domain_2019,bai_decaug_2021} is also an alternative, which can be seen as a relaxation of domain alignment, allowing partial features to be domain-specific and others to be domain-agnostic. In addition, some works put forward theoretical or experimental viewpoints from other perspectives. Instead of directly matching the representation distribution between source domains, Arjovsky et al. \cite{Arjovsky_Invariant_2019} force the classifier on the feature space to be optimal for all training environments, i.e., invariant risk minimization (IRM). SagNet \cite{NamLPYY21} reduce the domain gap by adjusting the model bias towards styles and contents to learn more semantically relevant representations.

In addition to data manipulation and representation learning, many works consider the DG problem from the perspective of \emph{model optimization}. Some methods improve the model robustness by combining general robust machine learning algorithms, including meta-learning \cite{li_learning_2018}, ensemble learning \cite{ding_deep_2018,seo_learning_2020}, adversarial learning \cite{GaninUAGLLML16,LiTGLLZT18}, and self-supervised learning \cite{KimYPKL21}. Besides, several works design regularization strategies based on intuition. RSC \cite{huang_selfChallenge_2020} avoids learning false low-level features by masking the feature components corresponding to large gradients to increase the difficulty of model training. GroupDRO \cite{sagawa2019distributionally} interlacing updates model parameters and the weights of each group. Groups with higher losses contribute more when updating model parameters. ANDMask \cite{ParascandoloNOG21} zeros out parameter gradient components with inconsistent signs across different environments during the model training based on gradient descent.


\textbf{Selected methods}. In this paper, we follow the DOMAINBED benchmark \cite{GulrajaniL21} and also incorporate some recent works, including eighteen algorithms chosen based on their impact over the years, their published performance, and a desire to include varied DG strategies mentioned above. A detailed list is shown in \mytabref{index}. Specifically, Mixup \cite{WangLK20} and SagNet \cite{NamLPYY21} are involved in data manipulation, and CORAL \cite{SunS16}, DANN \cite{GaninUAGLLML16}, CDANN \cite{LiTGLLZT18}, MMD \cite{LiPWK18}, IRM \cite{Arjovsky_Invariant_2019}, VREx \cite{KruegerCJ0BZPC21}, IB-ERM \cite{AhujaCZGBMR21}, IB-IRM \cite{AhujaCZGBMR21}, SelfReg \cite{KimYPKL21} and SagNet \cite{NamLPYY21} are related to representation learning. RSC \cite{HuangWXH20}, GroupDRO \cite{sagawa2019distributionally}, SelfReg \cite{KimYPKL21}, ANDMask \cite{ParascandoloNOG21}, SANDMask \cite{shahtalebi2021sand}, SD \cite{PezeshkiKBCPL21} and Fish \cite{ShiSTNHUS22} involve special considerations in model optimization.

\begin{table}[t]
\centering
\vspace{-1mm}
\caption{The eighteen DG methods we tested.}
\vspace{2mm}
\resizebox{1\textwidth}{!}{
\begin{tabular}{c|ccccccccc}
\hline
Index  & 1 & 2 & 3 & 4    & 5       & 6        & 7     & 8       & 9      \\ \hline
Method & CDANN \cite{LiTGLLZT18}     & DANN \cite{GaninUAGLLML16}        & IRM \cite{Arjovsky_Invariant_2019}       & Fish \cite{ShiSTNHUS22}    & VREx \cite{KruegerCJ0BZPC21}    & GroupDRO \cite{sagawa2019distributionally}    & ERM   & SelfReg \cite{KimYPKL21}    & SD \cite{PezeshkiKBCPL21}    \\ \hline
Index  & 10                        & 11                        & 12                        & 13   & 14      & 15       & 16    & 17      & 18     \\ \hline
Method & SANDMask \cite{shahtalebi2021sand}      & Mixup \cite{WangLK20}          & IB-IRM \cite{AhujaCZGBMR21}     & MMD \cite{LiPWK18}    & ANDMask \cite{ParascandoloNOG21}    & RSC \cite{HuangWXH20}     & CORAL \cite{SunS16}    & SagNet \cite{NamLPYY21}    & IB-ERM \cite{AhujaCZGBMR21}   \\ \hline
\end{tabular}}
\vspace{-1mm}
\label{index}
\end{table}

\subsection{Common Datasets}
Here is a brief look at six commonly used public datasets (PACS \cite{li2017deeper}, VLCS \cite{fang_unbiased_2013}, Colored MNIST \cite{Arjovsky_Invariant_2019}, DomainNet \cite{PengBXHSW19} and NICO \cite{he_towards_2021}) in the DG community. They are usually artificially divided into several domains, and the data in each domain has some consistency in visual form. All domains share the same category set.

\begin{figure}[H]
\begin{center}
\includegraphics[width=0.5\linewidth]{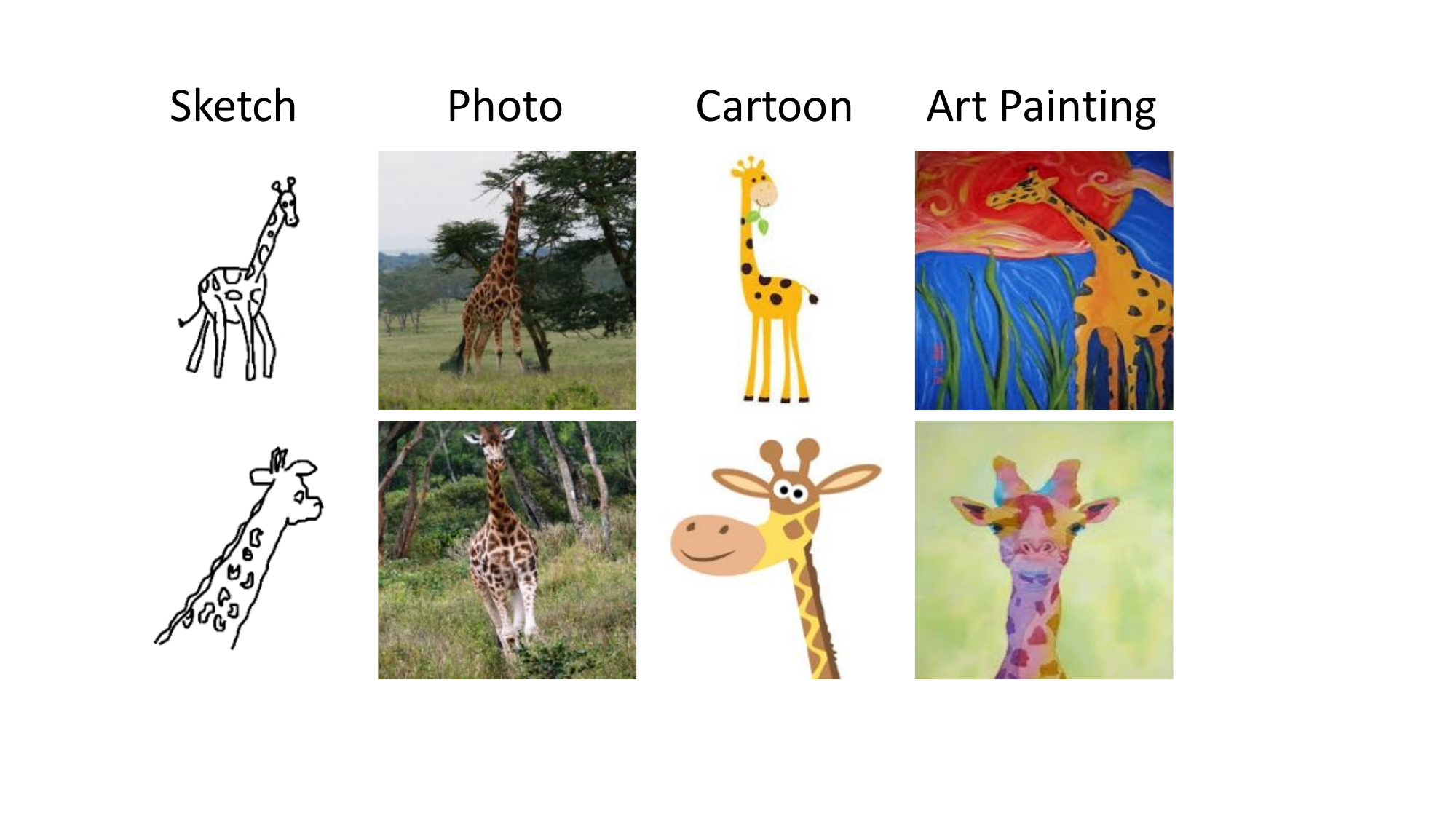}
\end{center}
\vspace{-6.5mm}
\caption{Example images of the ``giraffe" category from four domains of the PACS dataset.}
\label{PACS}
\end{figure}

\textbf{PACS} \cite{li2017deeper} consists of 7 categories (dog, elephant, giraffe, guitar, horse, house, person) across 4 different domains (Photo, Art Painting, Cartoon, Sketch) with totally 9991 images. It is a simple and balanced multi-domain dataset. All images have the same size $227 \times 227$. \myfigref{PACS} gives some examples of the dataset.

\begin{figure}[H]
\begin{center}
\includegraphics[width=0.5\linewidth]{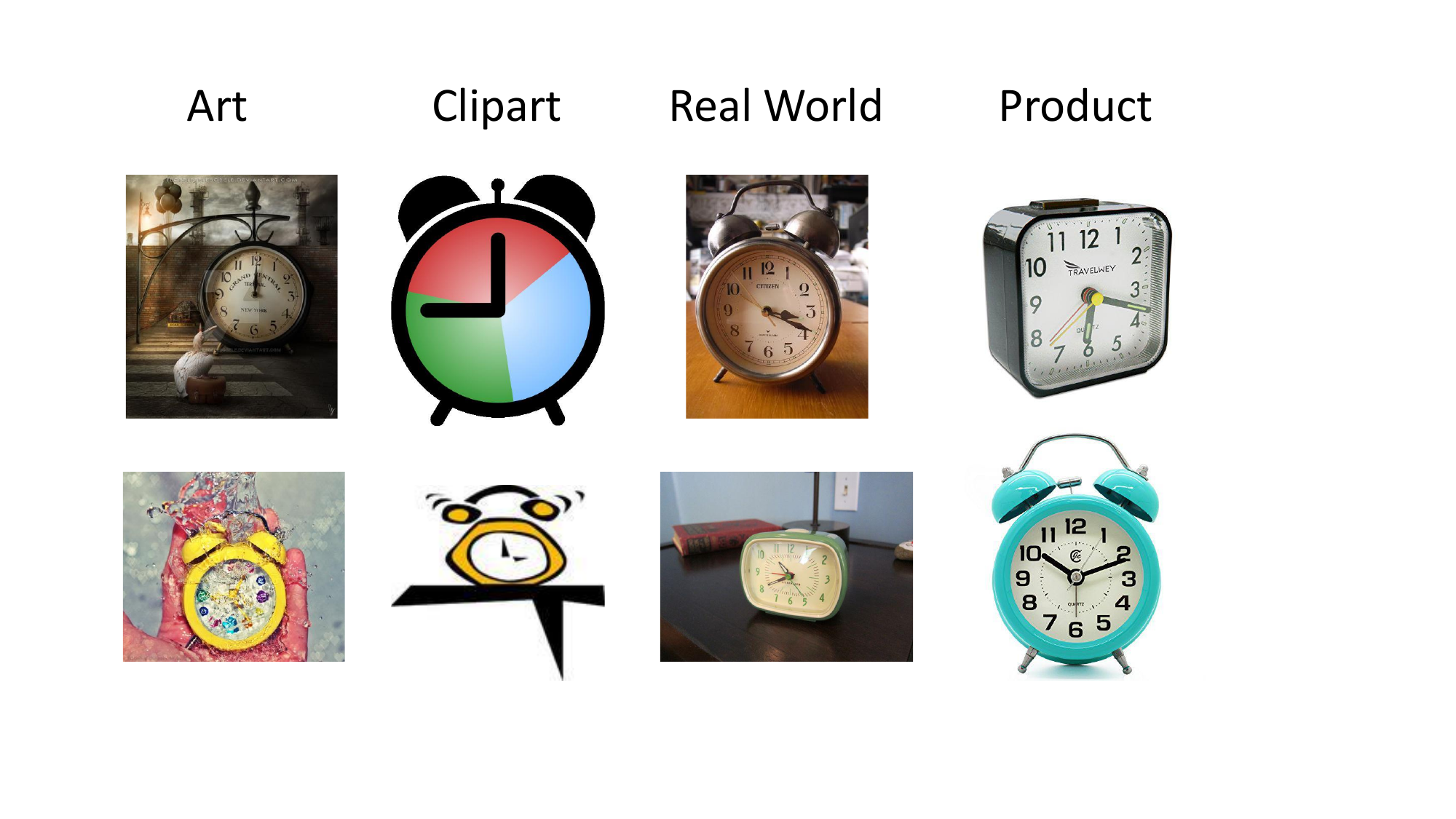}
\end{center}
\vspace{-6.5mm}
\caption{Example images of the ``alarm clock" category from four domains of the Office-Home dataset.}
\label{Office-Home}
\end{figure}

\textbf{Office-Home} \cite{venkateswara2017deep} includes 65 categories with totally 15588 images, also organized into 4 domains (Art, Clipart, Product, Real World). In particular, Art domain consists of artistic images in the form of sketches, paintings, ornamentation, etc. It contains objects typically appearing in office and home settings. Office-Home is a hard dataset with more subtle domain variations, larger categories and fewer samples per class than PACS. The dimensions of the images are not uniform. \myfigref{Office-Home} gives some examples from different domains of this dataset.

\begin{figure}[H]
\begin{center}
\includegraphics[width=0.5\linewidth]{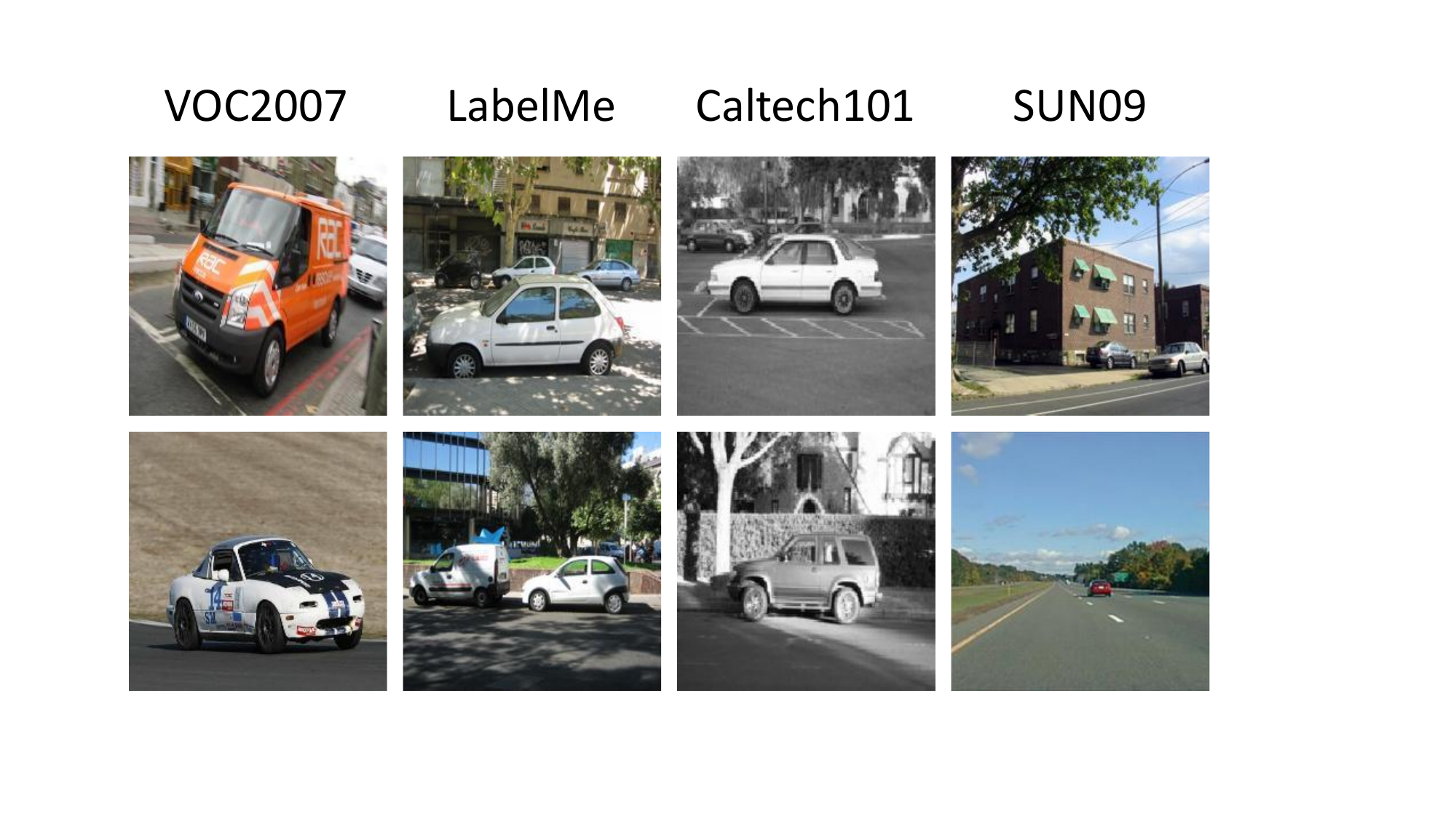}
\end{center}
\vspace{-6.5mm}
\caption{Example images of the ``vehicle" category from four domains of the VLCS dataset.}
\label{VLCS}
\end{figure}

\textbf{VLCS} \cite{fang_unbiased_2013} comprises five classes with 10729 images. It contains object-centric images from the Caltech-101 (C) dataset and scene-centric images from PASCAL VOC2007 (V), LabelMe (L), and SUN09 (S) datasets, each of which is seen as one domain. All images share the same size $227 \times 227$. \myfigref{VLCS} gives some examples of the VLCS dataset.

\begin{figure}[H]
\begin{center}
\includegraphics[width=0.82\linewidth]{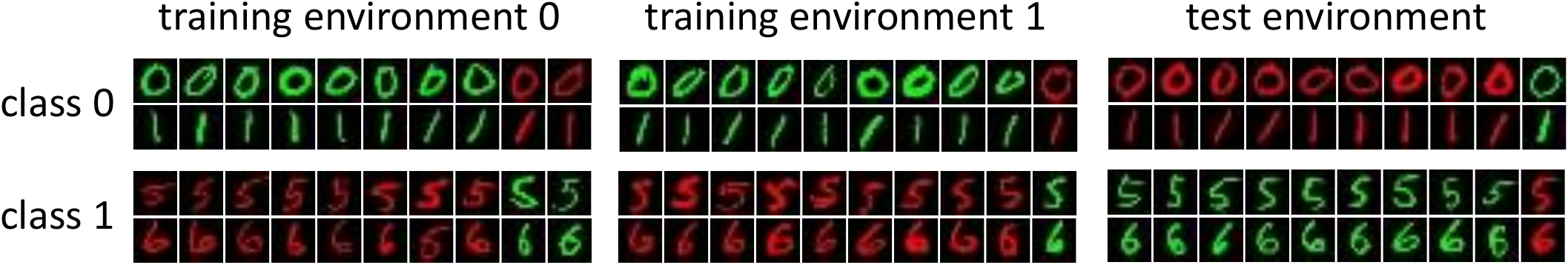}
\end{center}
\vspace{-6.5mm}
\caption{Example images of digits ``0, 1, 5, 6" from two training domains and one test domain in the Colored MNIST dataset. Digits 0 and 1 are labeled $\hat{y} = 0$, and digits 5 and 6 are labeled $\hat{y} = 1$. The final label $y$ is obtained by flipping $\hat{y}$ with probability 0.25.}
\label{Colored_MNIST}
\end{figure}

\textbf{Colored MNIST} \cite{Arjovsky_Invariant_2019} is a synthetic binary classification dataset ($\hat{y} = 0$ for digits 0-4 and $\hat{y} = 1$ for 5-9) derived from MINIST \cite{lecun1998gradient}. It associates class labels with red or green colors in the training set and reverses the correlation in the test set. This dataset is designed to study whether learning algorithms can avoid using spurious correlation features (in this case, color), which is harmful to generalization, for classification. \myfigref{Colored_MNIST} gives some examples of this dataset.

\begin{figure}[H]
\begin{center}
\includegraphics[width=0.7\linewidth]{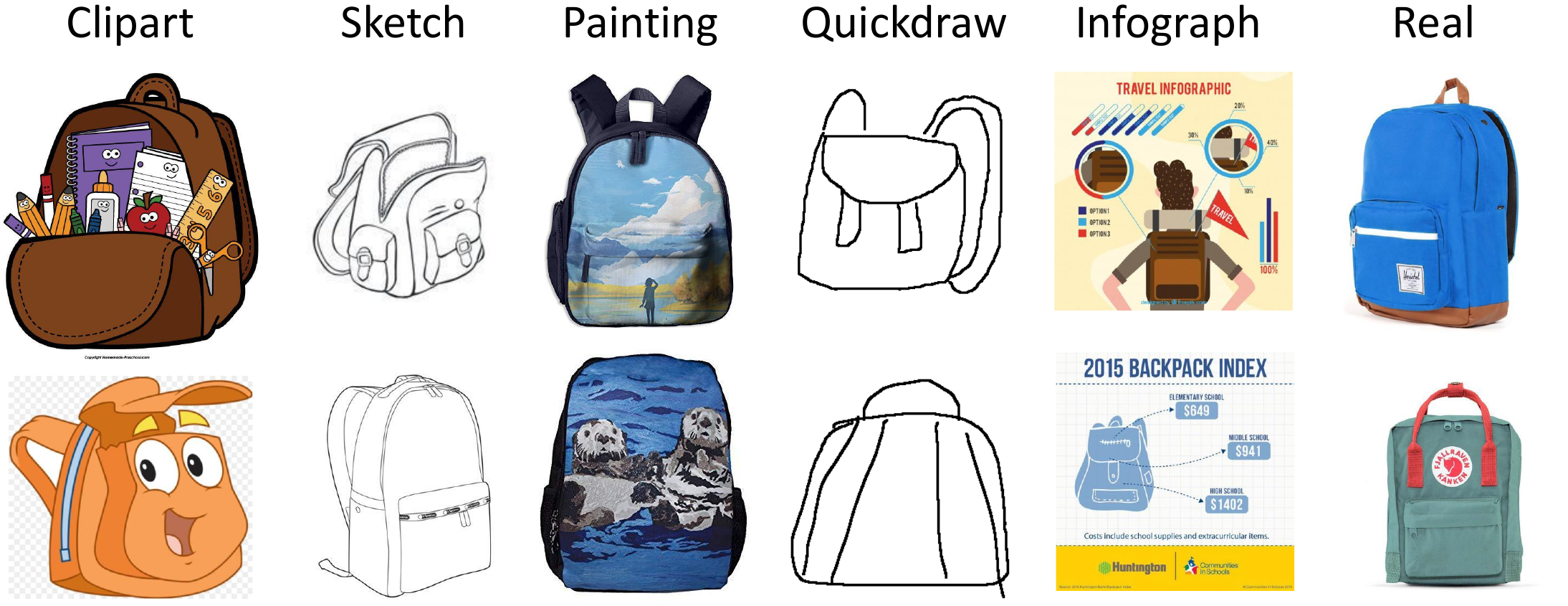}
\end{center}
\vspace{-6.5mm}
\caption{Example images of the ``backpack" category from six domains of the DomainNet dataset.}
\label{DomainNet}
\end{figure}

\textbf{DomainNet} \cite{PengBXHSW19} contaions 345 classes and 6 domains (clipart, infograph, painting, quickdraw, real, sketch). This dataset is a large-scale dataset with totally 586575 images. Except for the infograph domain, the sizes of different images vary slightly. \myfigref{DomainNet} gives some examples of this dataset.

\begin{figure}[H]
\begin{center}
\includegraphics[width=1\linewidth]{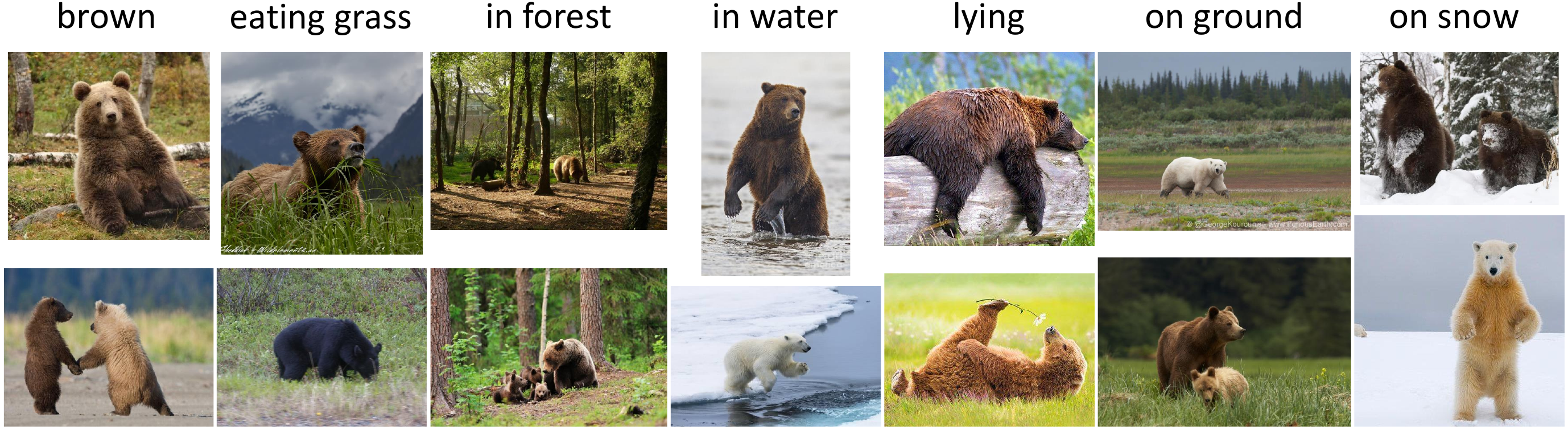}
\end{center}
\vspace{-6.5mm}
\caption{Example images of the ``bear" category from seven domains of the NICO dataset.}
\label{NICO}
\end{figure}

\textbf{NICO} \cite{he_towards_2021} consists of 19 classes from the animal and vehicle superclasses with nearly 25000 images in total. Within each class, images are divided into different contexts such as `on snow', `on grass', `aside people', `walking' and `lying'. One context can be seen as one domain. The sizes of images are not uniform. \myfigref{NICO} gives some examples of this dataset.

\subsection{Setting and Evaluation}
\label{evaluation}
Domain generalization aims to improve the generalization ability of a model to cope with out-of-distribution data. Existing domain generalization methods usually adopt the leave-one-domain-out protocol for evaluation. Specifically, for a dataset containing $n$ domains, $n-1$ domains are used as the sources while the remaining one serves as the test. In this way, there are $n-1$ selections in total. Under these selections, all test results and their average value are used as the performance evaluation. \mytabref{example_pacs} gives a more intuitive description.

\begin{table}[H]
\centering
\caption{An example of evaluations of three methods on the PACS dataset. `A', `C', `S' and `P' denote `Art painting', `Cartoon', `Sketch' and `Photo' domains, respectively. The cross-domain tasks are organized as `all source domains-test domain', for example, CPS-A.}
\vspace{2mm}
\resizebox{0.9\textwidth}{!}{
\begin{tabular}{ccccc|c}
\hline
    Methods & CPS-A & SPA-C & PAC-S & ACS-P & Average \\ \hline
    1 & $T_{1A}$ & $T_{1C}$ & $T_{1S}$ & $T_{1P}$ & $(T_{1A}+T_{1C}+T_{1S}+T_{1P})/4$ \\
    2 & $T_{2A}$ & $T_{2C}$ & $T_{2S}$ & $T_{2P}$ & $(T_{2A}+T_{2C}+T_{2S}+T_{2P})/4$ \\
    3 & $T_{3A}$ & $T_{3C}$ & $T_{3S}$ & $T_{3P}$ & $(T_{3A}+T_{3C}+T_{3S}+T_{3P})/4$ \\
\hline
\end{tabular}}
\label{example_pacs}
\end{table}

\section{A New Dataset: PaHCC}
As mentioned in Section \ref{DG_introduction}, there are many specialized domain generalization methods, which focus on improving the generalization ability of deep models in unseen domains by training on multiple available training domains (often termed as source domains). However, they are usually applied to object recognition tasks, and little research has been done on character recognition.

To enrich the relatively single application scenarios and visual patterns of existing public datasets, we propose a printed-to-handwritten Chinese character recognition task to facilitate the research on domain generalization methods. For the traditional handwritten Chinese character recognition task, researchers usually collect a large number of handwritten data to train the model, which is different from human cognition. Recall the way Chinese learn new characters, we first practice recognizing the printed fonts by analyzing the stroke structure of Chinese characters following the teacher, and then we naturally know the characters written by different people without too much practice. This shows that we humans have good domain generalization ability from printed fonts to handwritten characters. For the deep models, considering that we can generate amounts of synthetic printed data utilizing font libraries easily, it is of great significance if we can train a robust recognition model that performs well on handwritten characters from different writers, using only synthetic printed data.

For this challenging practical task, we construct a large-scale classification dataset of \textbf{P}rinted \textbf{a}nd \textbf{H}andwritten \textbf{C}hinese \textbf{C}haracters (PaHCC). We will make the PaHCC dataset public after the publishing of this paper. Below we describe the dataset in detail.

\subsection{Database Creation Process}
Here we selected 1000 frequently used Chinese characters in the GB2312-80 standard as our categories. The synthetic printed Chinese character images decode from the SCUT-SPCCI database that contains synthetic printed character images generated from 280 different fonts (\url{http://www.hcii-lab.net/data/scutspcci/download.html}). All images are gray images of size $64\times64$. \myfigref{printed_d0}, \myfigref{printed_d1} and \myfigref{printed_d2} show some representative examples. For handwritten Chinese characters, we decode images from CASIA-HWDB1.0-1.1 \cite{LiuYWW11} that contains gray images segmented and labeled from scanned handwritten pages. We preserve the original relative sizes of images, and \myfigref{handwritten_t} shows some examples.

\subsection{Data Structure and Statistics}
\begin{figure}[t]
\begin{center}
\subfigure[standard printed fonts]{
\includegraphics[width=0.47\columnwidth]{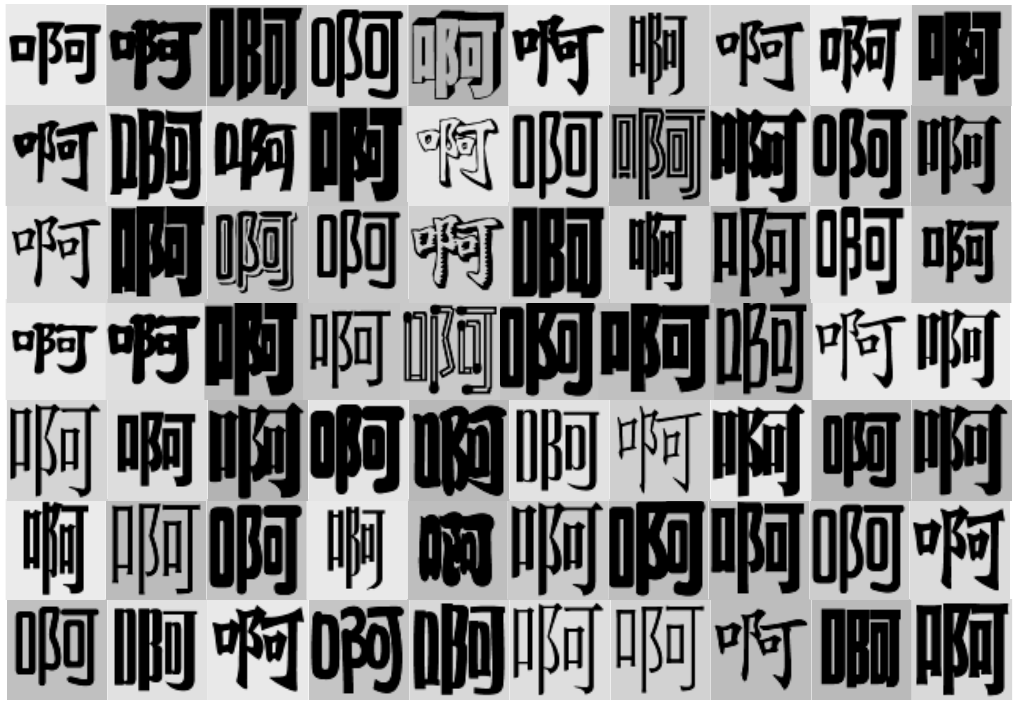}
\label{printed_d0}
}
\subfigure[distorted printed fonts]{
\includegraphics[width=0.47\columnwidth]{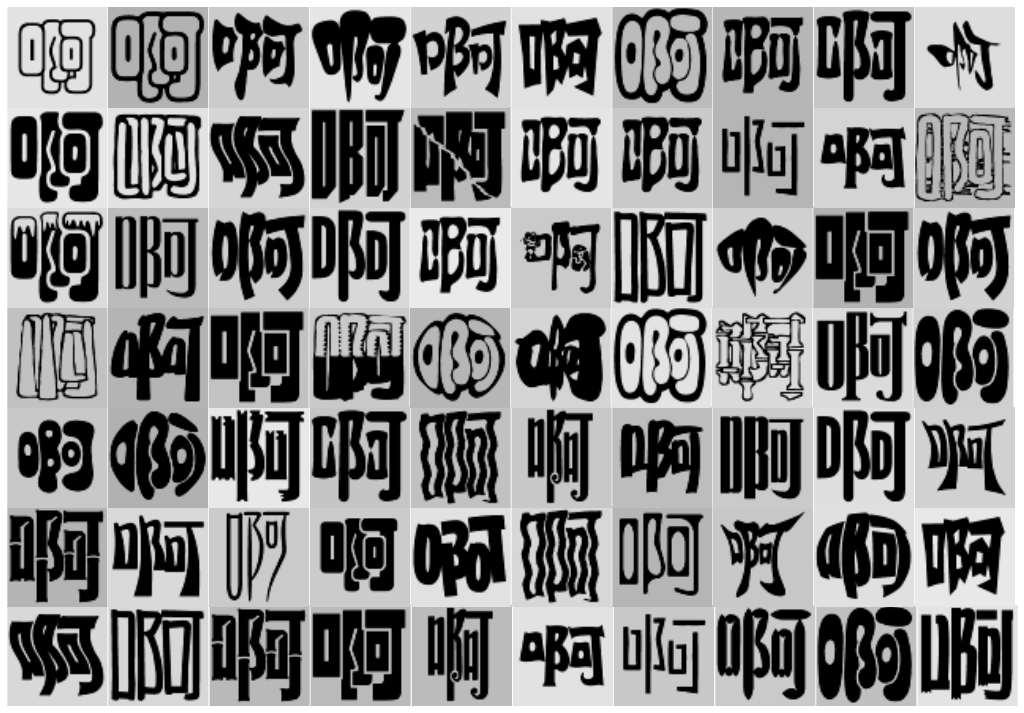}
\label{printed_d1}
}
\subfigure[handwriting-style printed fonts]{
\includegraphics[width=0.47\columnwidth]{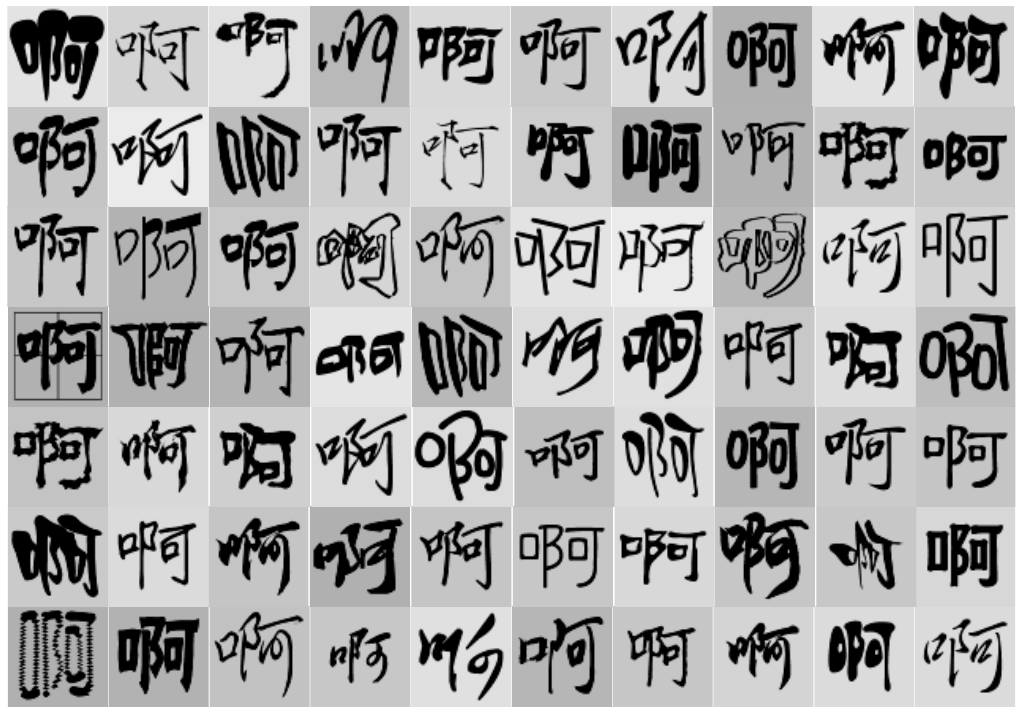}
\label{printed_d2}
}
\subfigure[scanned handwritten data]{
\includegraphics[width=0.47\columnwidth]{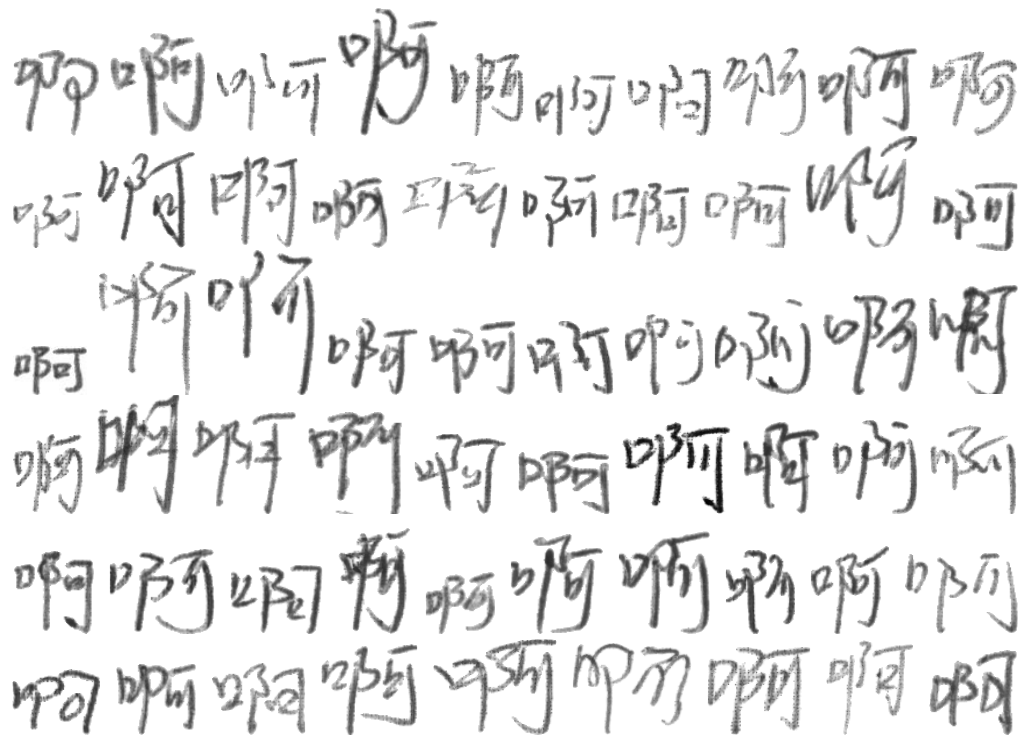}
\label{handwritten_t}
}
\caption{Examples of the PaHCC dataset. (a)-(c) are synthetic printed Chinese character images. (d) is scanned handwritten Chinese character images.}
\label{example_images}
\end{center}
\end{figure}

\textbf{PaHCC} is our full dataset, which contains 1000 classes and 996478 samples. The printed data comprises 280647 synthetic Chinese character images with about 280 printed fonts, divided into three domains according to font types (standard printed fonts, distorted printed fonts, and handwriting-style printed fonts). The handwritten data contains 715831 scanned handwritten Chinese character images from 720 writers. \myfigref{example_images} provides a visual representation of our data structure. We present more fine-grained statistics of the dataset in \mytabref{big_train} and \mytabref{big_test}.

\begin{table}[H]
\centering
\caption{Statistics of the synthetic printed data in our PaHCC dataset.}
\vspace{2mm}
\begin{tabular}{l|c|c|l}
\hline
\multicolumn{1}{c|}{} & \#fonts & \#samples per class & \multicolumn{1}{c}{\#samples} \\ \hline
Domain 0              & 94                              & $\sim$94                                    & 94229                                                 \\
Domain 1              & 93                              & $\sim$93                                    & 93238                                                 \\
Domain 2              & 93                              & $\sim$93                                    & 93180                                                 \\ \hline
Total                 & 280                             & $\sim$280                                   & 280647                                                \\ \hline
\end{tabular}
\label{big_train}
\end{table}

\begin{table}[H]
\centering
\caption{Statistics of the scanned handwritten data in our PaHCC dataset.}
\vspace{2mm}
\begin{tabular}{l|c|c|l}
\hline
\multicolumn{1}{c|}{} & \#fonts & \#samples per class & \multicolumn{1}{c}{\#samples} \\ \hline
Domain 0              & 720                              & $\sim$716                                    & 715831                                                 \\ \hline
\end{tabular}
\label{big_test}
\end{table}

Considering the large scale of our complete dataset (PaHCC), we also organize a small dataset of 100 classes (\textbf{mini-PaHCC}) to ease the computational cost of the study. Our main experiments are also based on the small dataset. \mytabref{mini_train} and \mytabref{mini_test} present fine-grained statistics of our mini-PaHCC dataset.

\begin{table}[H]
\vskip -0.1in
\centering
\caption{Statistics of the synthetic printed data in our mini-PaHCC dataset.}
\vspace{2mm}
\begin{tabular}{l|c|c|l}
\hline
\multicolumn{1}{c|}{} & \#fonts & \#samples per class & \multicolumn{1}{c}{\#samples} \\ \hline
Domain 0              & 94                              & $\sim$94                                    & 9428                                                 \\
Domain 1              & 93                              & $\sim$93                                    & 9324                                                 \\
Domain 2              & 93                              & $\sim$93                                    & 9327                                                 \\ \hline
Total                 & 280                             & $\sim$280                                   & 28079                                                \\ \hline
\end{tabular}
\label{mini_train}
\end{table}

\begin{table}[H]
\vskip -0.2in
\centering
\caption{Statistics of the scanned handwritten data in our mini-PaHCC dataset.}
\vspace{2mm}
\begin{tabular}{l|c|c|l}
\hline
\multicolumn{1}{c|}{} & \#fonts & \#samples per class & \multicolumn{1}{c}{\#samples} \\ \hline
Domain 0              & 720                              & $\sim$717                                    & 71659                                                 \\ \hline
\end{tabular}
\label{mini_test}
\end{table}

\subsection{Challenging Tasks}
Our PaHCC is a large-scale and comprehensive Chinese character dataset. It can support research on many challenging problems related to the model's robustness, transferability, and interpretability in visual pattern recognition. We give recommendations as follows.

\emph{1) Domain Generalization:} Our dataset can naturally support the study of domain generalization, which focuses on improving the generalization ability of predictive models in unseen domains by training on multiple available source domains. A practical setting is to use synthetic printed data for model training and handwritten data for testing, which minimizes the data collection cost for the handwritten Chinese character recognition task.

\emph{2) Domain Adaptation:} Similar to domain generalization, PaHCC can also be used to evaluate domain adaptation methods when we allow the distribution of the test data to be available. In addition, it is also an alternative to use only synthetic printed data and evaluate the robustness of the model to printed fonts by leaving one part as the test set, the same for the handwritten data.

\emph{3) Structure-understanding Model:} PaHCC contains a wide variety of styles and forms, which strongly challenges the robustness of deep models. However, Chinese characters have unique structure information. Following the way humans learn and perceive, it is promising for a deep model to achieve stable and excellent recognition ability utilizing this structural information. Therefore, PaHCC is very suited for developing interpretable models with structure understanding.

\emph{4) Zero-Shot Learning:} For the handwritten Chinese character recognition task, PaHCC can facilitate the research on zero-shot learning by utilizing readily available synthetic printed data as auxiliary information.

\emph{5) Class-Incremental Learning:} Since PaHCC is a large-scale dataset with 1000 classes, it is convenient to divide the data of a domain along categories to assist the research on class-incremental learning.

\section{Experiments}
For all experiments, we use the open-source implementation (\url{https://github.com/facebookresearch/DomainBed}) of the DG benchmark released by the authors of DomainBed \cite{GulrajaniL21}. On our Chinese characters datasets (PaHCC and mini-PaHCC), due to the small size of the character image, we use the ResNet-18 backbone and modify the kernel size of conv1 from default $7\times7$ to $3\times3$. On the DomainNet dataset, we follow the default configuration in DomainBed.

\subsection{Existing DG methods cannot work well on the PaHCC dataset.}
We first investigated the performance of ERM on the mini-PaHCC dataset. When training the predictive model, we only use the classification loss of cross-entropy. 

According to the experimental results on the original grayscale images in \mytabref{ERM_mini_PaHCC}, the model trained with ERM has good generalization ability (90\%+) between different printed domains. However, when the model is transferred from printed characters to handwritten ones, the performance decreases sharply (from 90\%+ to 16\%). By observing the dataset, we find that on the original grayscale images, the foreground pixel value of handwritten data is very uneven, and it is generally uniform in synthesized printed data. For the Chinese character recognition task, this statistical difference in underlying pixels plays a significant impact on model performance. It indicates that the deep model trained with ERM mainly uses statistics of pixels rather than structural information of characters to classify, which is a shortcut learning phenomenon. After we binarize all the images, the generalization between different printed fonts does not have much effect, while the printed-to-handwritten task is greatly improved (from 16\% to 60\%), which verifies our above analysis. To alleviate this problem, researchers usually adopt some necessary preprocessing operations \cite{LiuYWW13}. But this does not cure the problems of the learning algorithm and the model itself. 

\begin{table}[H]
\vskip -0.1in
\centering
\caption{DG accuracies of ERM on the mini-PaHCC dataset under different settings. ``d0", ``d1" and ``d2" represent domain 0, domain 1, and domain 2 in \mytabref{mini_train}, respectively, corresponding to \myfigref{printed_d0}, \myfigref{printed_d1}, and \myfigref{printed_d2}. ``handwritten" represents the scanned handwritten data in \mytabref{mini_test}, corresponding to \myfigref{handwritten_t}. The accuracy is achieved by \emph{training-domain validation} \cite{GulrajaniL21} as the model selection method.}
\vspace{2mm}
\begin{tabular}{c|cc}
\hline
\multicolumn{1}{c|}{Task} & grayscale image   & binary image \\ \hline
printed ($d_0, d_1$) $\rightarrow$ printed ($d_2$)          & 90.1 $\pm$ 0.7 & 87.2 $\pm$ 0.5 \\
printed ($d_0, d_2$) $\rightarrow$ printed ($d_1$)          & 91.8 $\pm$ 0.6 & 91.2 $\pm$ 0.1 \\
printed ($d_1, d_2$) $\rightarrow$ printed ($d_0$)          & 99.6 $\pm$ 0.0 & 99.5 $\pm$ 0.1 \\ \hline
($d_0, d_1, d_2$) $\rightarrow$ handwritten         & 16.4 $\pm$ 4.7 & \bf{60.6 $\pm$ 0.7} \\ \hline
\end{tabular}
\label{ERM_mini_PaHCC}
\vskip -0.1in
\end{table}
 
\begin{figure}[t]
\begin{center}
\includegraphics[width=0.95\linewidth]{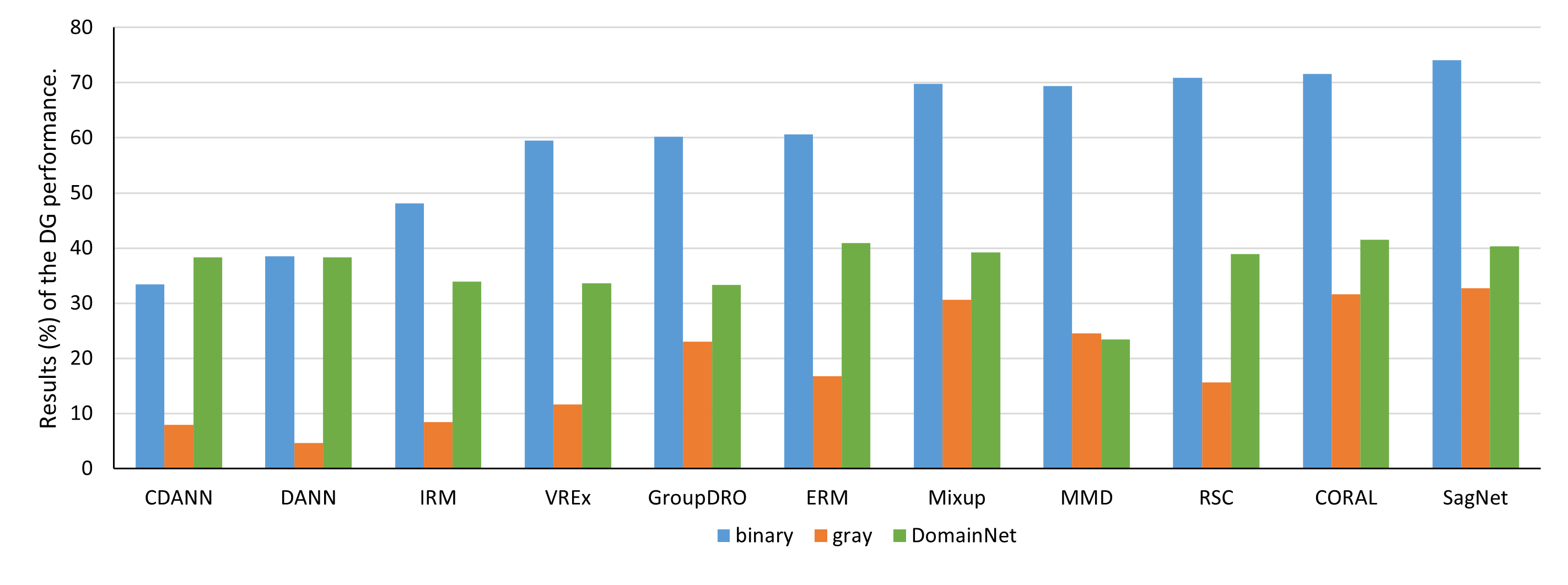}
\end{center}
\vspace{-9mm}
\caption{The performance comparison of various methods on the mini-PaHCC and DomainNet datasets. Blue bars are the results of binary images on the mini-PaHCC dataset. Orange bars are the results of the original gray images on the mini-PaHCC dataset. Green bars are the results on the DomainNet datasets. The accuracy is achieved by \emph{training-domain validation} \cite{GulrajaniL21} as the model selection method.}
\vspace{-2mm}
\label{PaHCC_0}
\end{figure}

Further, we investigate the performance of multiple DG methods on original gray images and binary images. As is shown in \myfigref{PaHCC_0}, The handwritten generalization performances of all the methods on the original gray data (orange bars) are very poor. After all the data are processed into binary images (blue bars), the performances are greatly improved ($+26\% \sim +55\%$), which shows the same phenomenon as ERM. It indicates that these specialized domain generalization methods do not ease the model's shortcut learning of pixel statistics. In particular, the results of these DG methods on the 100-class mini-PaHCC dataset (orange bars) are significantly lower than those on the 345-class DomainNet dataset (green bars), demonstrating a harder domain generalization challenge.

The detailed performance of more algorithms on the mini-PaHCC and PaHCC datasets is shown in \mytabref{results_PaHCCandMiniPaHCC}. Since the results on original grayscale images are too low, see orange bars in \myfigref{PaHCC_0} for details, we only conduct subsequent experiments on binary images. On both datasets, we conduct hyperparameters search as Domainbed \cite{GulrajaniL21} (a random search of 20 trials over a joint distribution of all hyperparameters), and report a mean (and its standard error) over three repetitions. The detailed performance on the mini-PaHCC and PaHCC datasets is shown in \mytabref{results_PaHCCandMiniPaHCC}.

\begin{table}[t]
\centering
\caption{Performances of different algorithms on the mini-PaHCC and PaHCC datasets with binary images. Accuracies are achieved by \emph{training-domain validation} \cite{GulrajaniL21} as the model selection methods.}
\vspace{2mm}
\begin{tabular}{l|c|c}
\hline
\multicolumn{1}{l|}{Algorithm}  & mini-PaHCC (100 classes) & PaHCC (1000 classes) \\ \hline
ERM          & 60.6 $\pm$ 0.7 & 63.7 $\pm$ 1.7 \\
CDANN \cite{LiTGLLZT18}        & 33.4 $\pm$ 2.6 & - \\
DANN \cite{GaninUAGLLML16}        & 38.5 $\pm$ 2.1 & - \\
IRM \cite{Arjovsky_Invariant_2019}        & 48.1 $\pm$ 10.6 & - \\
Fish \cite{ShiSTNHUS22}        & 56.5 $\pm$ 4.0 & - \\
VREx \cite{KruegerCJ0BZPC21}        & 59.5 $\pm$ 2.6 & - \\
GroupDRO \cite{sagawa2019distributionally}        & 60.2 $\pm$ 3.9 & - \\
SelfReg \cite{KimYPKL21}        & 60.6 $\pm$ 7.2 & - \\
SD \cite{PezeshkiKBCPL21}        & 64.2 $\pm$ 4.6 & - \\
SANDMask \cite{shahtalebi2021sand}        & 64.7 $\pm$ 1.9 & - \\
IB-IRM \cite{AhujaCZGBMR21}        & 67.0 $\pm$ 6.0 & - \\
MMD \cite{LiPWK18}        & 69.4 $\pm$ 3.6 & - \\
ANDMask \cite{ParascandoloNOG21}        & 69.5 $\pm$ 2.8 & 61.0 $\pm$ 1.6  \\
Mixup \cite{WangLK20}        & 69.8 $\pm$ 1.2 & - \\
RSC \cite{HuangWXH20}        & 70.9 $\pm$ 1.8 & 63.0 $\pm$ 1.3 \\
CORAL \cite{SunS16}        & 71.6 $\pm$ 1.1 & 59.7 $\pm$ 0.2 \\
SagNet \cite{NamLPYY21}        & 74.1 $\pm$ 1.2 & 67.7 $\pm$ 0.7 \\
IB-ERM \cite{AhujaCZGBMR21}        & \bf{77.3 $\pm$ 0.5} & 71.7 $\pm$ 1.8 \\ \hline
\end{tabular}
\label{results_PaHCCandMiniPaHCC}
\end{table}

On the mini-PaHCC dataset, some of the tested eighteen DG algorithms significantly improve the performance compared to ERM, such as the representation learning methods MMD ($+9\%$), CORAL ($+11\%$), SagNet ($+13.5\%$), and IB-ERM ($+17\%$), the learning strategy methods ANDMask ($+9\%$) and RSC ($+10\%$), and the data augmentation method Mixup ($+9\%$). However, some works (e.g., CDANN, DANN, and IRM) deteriorate the performance, which is consistent with the observation of DomainBed \cite{GulrajaniL21} on seven publicly available object recognition datasets. In particular, IRM and VREx, methods with theoretical guarantees fail in practice. It is generally accepted that overparameterization is the main reason for the failure of these methods on deep models and large datasets \cite{ZhouLZZ22}.

On the PaHCC dataset, a large-scale dataset of 1000 classes, performances of all tested methods show a slight decline, as expected. It is worth noting that some methods, such as ANDMask, RSC and CORAL, cannot maintain the original effects on the large category dataset. By contrast, SagNet and IB-ERM, the two representation learning methods, consistently achieve superior performance. However, the optimal performances on both datasets are still less than 80\%, indicating that existing DG methods do not solve our task well, and there is still much room for improvement.

\subsection{Unreliable Evaluation under the Standard DG Setting}
\label{evaluation_experiment}
In previous studies, researchers paid little attention to the selection of training domains. The following questions are still not well investigated: \emph{How many training domains and what kinds of training domains would be more conducive to training a model with good generalization ability?} In this section, through a series of ``dynamic" experiments, we demonstrate that it is unreliable to use the commonly used leave-one-domain-out protocol (see Section \ref{evaluation} for more details) under the standard DG setting for algorithm evaluation. Next, we will elaborate on two important observations.

\subsubsection{The selection of training domains affects the performance of DG methods without changing the number of training domains.}

\begin{figure}[t]
\begin{center}
\includegraphics[width=0.65\linewidth]{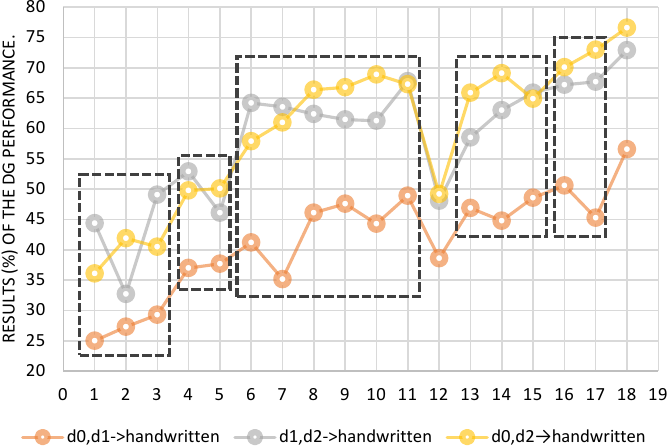}
\end{center}
\vspace{-6.5mm}
\caption{The performance comparison of various methods on the mini-PaHCC dataset. The DG methods corresponding to the indices on the horizontal axis are shown in \mytabref{index}. The accuracy is achieved by \emph{training-domain validation} \cite{GulrajaniL21} as the model selection method. For details about the meaning of ``d0", ``d1", ``d2" and ``handwritten", see \mytabref{ERM_mini_PaHCC}.}
\label{PaHCC}
\vskip -0.1in
\end{figure}

\begin{figure}[H]
\begin{center}
\subfigure{
\includegraphics[width=0.65\columnwidth]{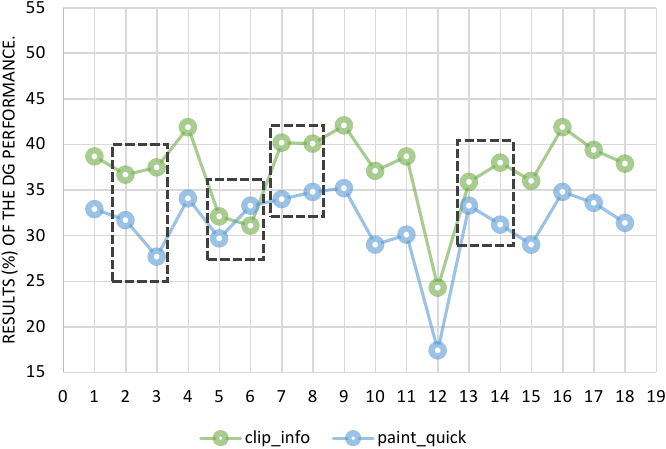}
\label{DomainNet_0_0}
}
\subfigure{
\includegraphics[width=0.65\columnwidth]{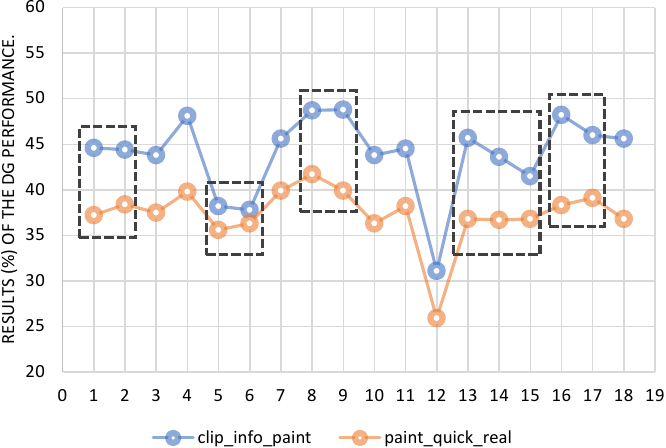}
\label{DomainNet_0_1}
}
\subfigure{
\includegraphics[width=0.65\columnwidth]{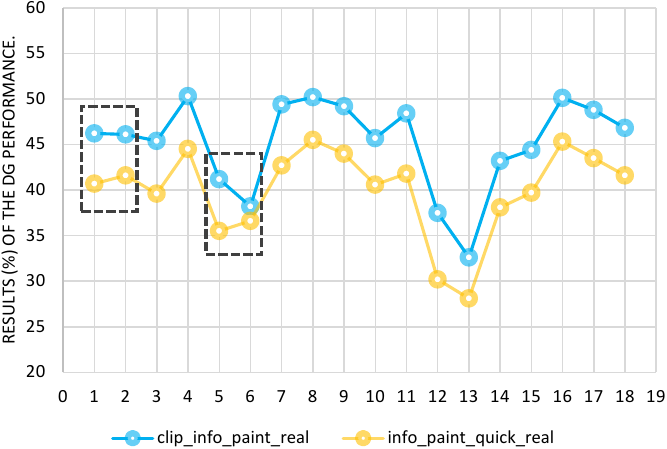}
\label{DomainNet_0_2}
}
\vskip -0.07in
\caption{The performance comparison of various methods on the DomainNet dataset. The DG methods corresponding to the indices on the horizontal axis are shown in \mytabref{index}. The accuracy is achieved by \emph{training-domain validation} \cite{GulrajaniL21} as the model selection method. ``clip", ``info", ``paint", ``quick" and ``real" represent domains ``clipart", ``infograph", ``painting", ``quickdraw" and ``real", respectively. All models are tested on the ``sketch" domain. Visual examples of each domain are shown in \myfigref{DomainNet}.}
\label{DomainNet_0}
\end{center}
\end{figure}

Here we investigate the impact of different training domain choices on the performance of DG methods when the number of training domains is unaltered. We consider training models with different source domains and testing on the same unseen one. We conducted experiments on the mini-PaHCC dataset (Chinese character recognition) and the DomainNet dataset (object recognition), respectively. \myfigref{PaHCC} shows the performance comparison of various methods on the mini-PaHCC dataset. Three selections of source domains were considered: d0 and d1 (the orange line), d1 and d2 (the gray line), and d0 and d2 (the yellow line). According to the accuracies of the handwritten data under the three settings, we find a common phenomenon that although the method $a$ is better than the method $b$ under one setting, the relationship could reverse if the selected training domains are changed. \myfigref{DomainNet_0} shows the performance comparison of various methods on the DomainNet dataset. It also indicates that the effects of DG methods may change with different selections of training domains, but not as clearly as in mini-PaHCC.

\subsubsection{The performance of some DG methods deteriorates when increasing the number of training domains.}
Here, we consider adding new source domain data to the original ones for training and observe its impact on different methods before and after the addition. All experiments in this section are also based on the DomainBed codebase. 

\begin{figure}[H]
\begin{center}
\subfigure[Adding a new training domain (d2: \myfigref{printed_d2}) whose style is close to the test domain.]{
\includegraphics[width=1\columnwidth]{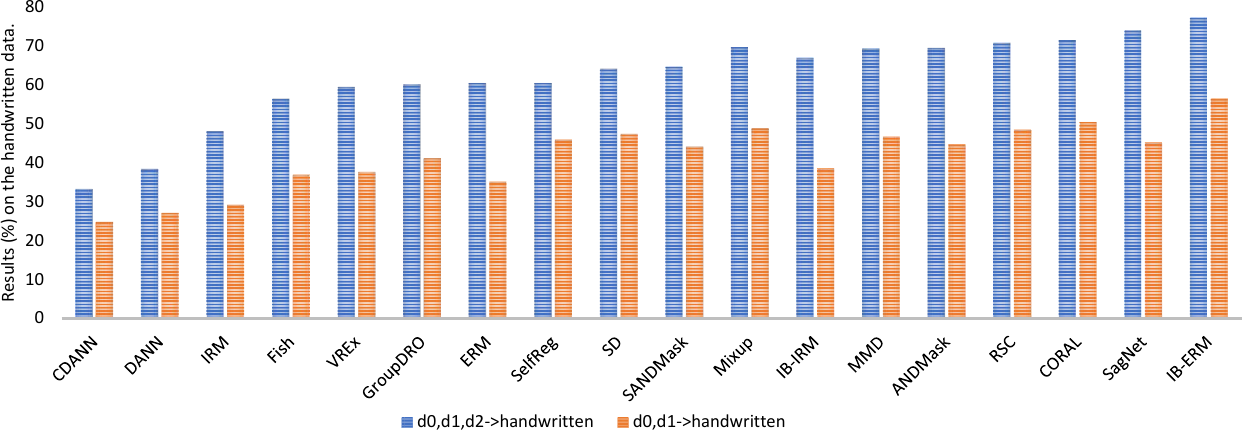}
\label{PaHCC_1}
}
\subfigure[Adding a new training domain (d0: \myfigref{printed_d0}) whose style is quite different from the test domain.]{
\includegraphics[width=1\columnwidth]{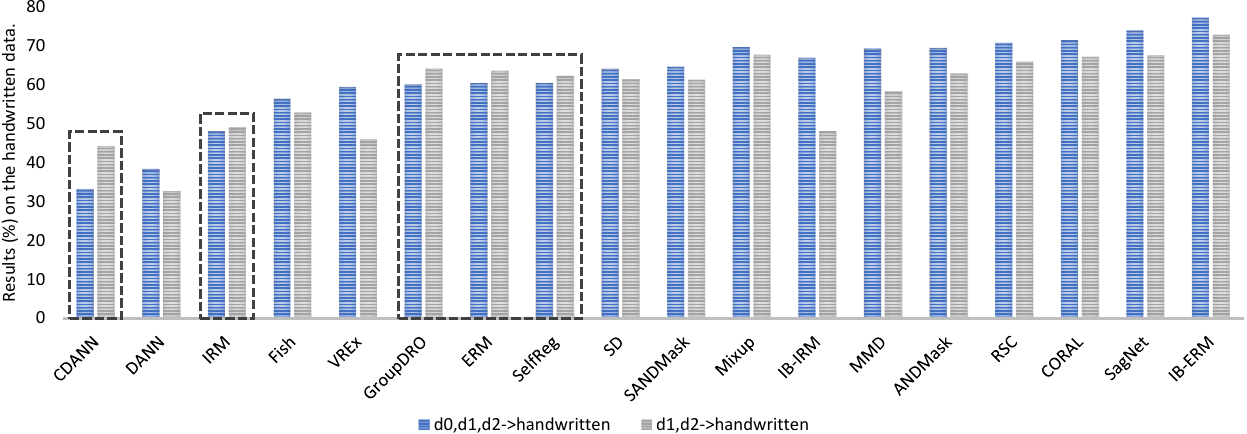}
\label{PaHCC_2}
}
\subfigure[Adding a new training domain (d1: \myfigref{printed_d1}) whose style is quite different from the test domain.]{
\includegraphics[width=1\columnwidth]{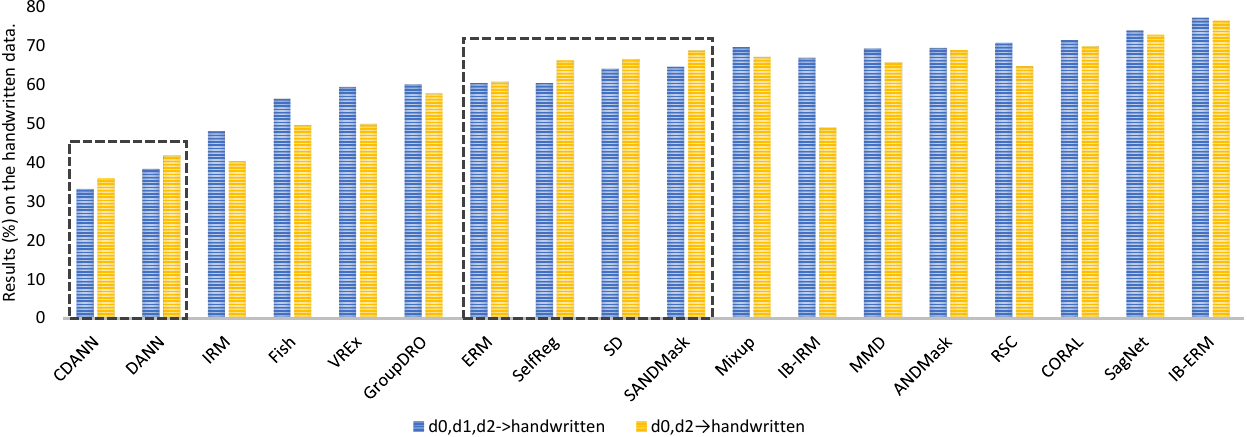}
\label{PaHCC_3}
}
\vskip -0.07in
\caption{The performance comparison of various methods on the mini-PaHCC dataset. The accuracy is achieved by \emph{training-domain validation} \cite{GulrajaniL21} as the model selection method. For details about the meaning of ``d0", ``d1", ``d2" and ``handwritten", see \mytabref{ERM_mini_PaHCC}.}
\label{PaHCC_increase}
\end{center}
\end{figure}

On the mini-PaHCC dataset, we consider the generalization ability of a model trained on synthetic printed Chinese characters to scanned handwritten ones. Since our PaHCC dataset only has three printed domains, we consider adding the remaining printed domain to the three settings in \myfigref{PaHCC} respectively and observe the responses of different methods. The experimental results are shown in \myfigref{PaHCC_1}, \myfigref{PaHCC_2}, and \myfigref{PaHCC_3}, respectively. \myfigref{PaHCC_1} shows that all DG methods and ERM can achieve better results when adding a new training domain (d2: \myfigref{printed_d2}) whose style is close to the test domain. \myfigref{PaHCC_2} and \myfigref{PaHCC_3} show that the performances of some DG methods (and, of course, ERM) deteriorate when adding a new training domain (d0: \myfigref{printed_d0}; d1: \myfigref{printed_d1}) whose style is quite different from the test domain, which indicates that these methods cannot cope well with distribution shifts. 

\begin{figure}[H]
\begin{center}
\subfigure[Adding new training domains continuously on the base training domains ``painting" and ``quickdraw".]{
\includegraphics[width=1\columnwidth]{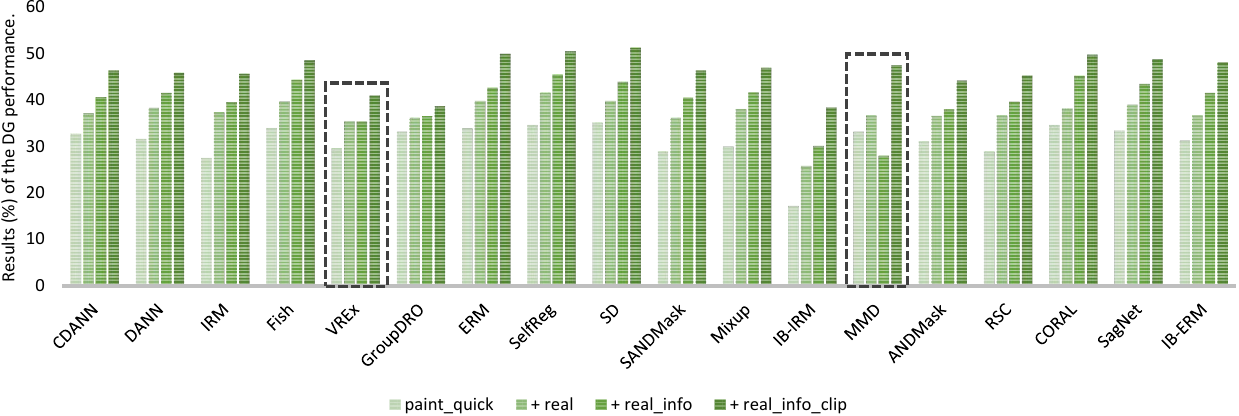}
\label{DomainNet_1}
}
\subfigure[Adding new training domains continuously on the base training domains ``clipart" and ``infograph".]{
\includegraphics[width=1\columnwidth]{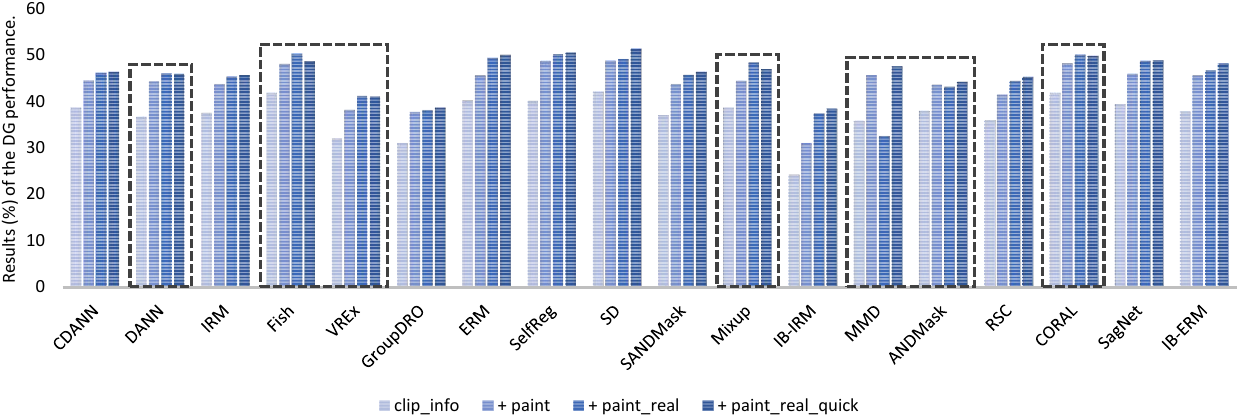}
\label{DomainNet_2}
}
\vskip -0.07in
\caption{The performance comparison of various methods on the DomainNet dataset. The accuracy is achieved by \emph{training-domain validation} \cite{GulrajaniL21} as the model selection method. For details about the meaning of ```clip", ``info", ``paint", ``quick" and ``real", see \myfigref{DomainNet_0}. All models are tested on the ``sketch" domain.}
\label{DomainNet_increase}
\end{center}
\end{figure}

For the DomainNet dataset, we have five training domains to choose from under the leave-one-domain-out protocol. We continuously increase the number of training domains for model training and observe the domain generalization performances of different methods. We consider two base cases (training domains: painting and quickdraw; clipart and infograph) and perform three rounds of adding new training domains, respectively. Add one new training domain each time. All models are tested on the ``sketch" domain. The experimental results are shown in \myfigref{DomainNet_increase}. In both cases, some DG methods exhibit performance degradation as the training domain increases.

\subsection{More Reliable Evaluations under the Dynamic DG Setting}
In Section \ref{evaluation_experiment}, we show that the evaluation of DG methods under the standard setting is unreliable from following two aspects. 1) The selection of training domains under the unaltered number of them affects the performance of DG methods. 2) Increasing the number of training domains may degrade the performance of some DG methods. Recent empirical benchmarks \cite{GulrajaniL21,KohSMXZBHYPGLDS21} also show that although existing dedicated DG methods are technically sound, their effectiveness under a unified experimental condition has been strongly challenged by ERM. To promote substantial progress in domain generalization, we advocate that researchers in this community can refer to the performance of DG methods under the dynamic DG setting for more reliable evaluation. For the dynamic DG setting, we will expand it in detail below.

As our evaluations in Section \ref{evaluation_experiment} show, in this paper, we consider the performance of the DG method under two dynamic forms: \emph{performance when increasing the number of training domains}; \emph{performance when selecting different training domains (the number is unaltered)}. That's not the only thing fixed, and there may be more considerations in the future. The dynamic DG setting can assist the common leave-one-domain-out protocol (static) with the method evaluation comprehensively and reliably. Specifically, if a method can improve the performance on unseen domain data after adding new training domains whose distribution is quite different from the test domain, the method can effectively utilize more source domain data to help the model cope with domain shifts better. We say that this method is capable to cope with distribution shifts. We call it the ``\textbf{Domain +}" evaluation criterion. In addition, if one method's advantage compared with other methods does not reverse when choosing different training domains without changing the number of training domains, this method has a stable performance advantage. We call it the ``\textbf{ST (swap test)}" evaluation criterion.

Under the ``Domain +" evaluation criterion, we have the following findings. 1) The failed methods under the ``Domain +" criterion on the PaHCC and DomainNet datasets are quite different. It may be due to the large gap between the types of domain shifts in Chinese character recognition and object recognition. Therefore, it is better to test on as many different recognition tasks as possible to evaluate the generality of the algorithm. 2) ERM, which exhibits excellent performance compared to numerous DG methods on object recognition datasets, suffers performance degradation on the Chinese character recognition dataset after adding new training domain data whose style is quite different from the test domain (see \myfigref{PaHCC_2} and \myfigref{PaHCC_3}), indicating that it don't cope with the distribution shifts well. 3) According to the ``Domain +" criteria, some of these failed methods under the original leave-one-domain-out protocol, such as IB-IRM (see \myfigref{PaHCC_2}, \myfigref{PaHCC_3}, \myfigref{DomainNet_1} and \myfigref{DomainNet_2}), still can cope with distribution shifts. They can achieve better performance than ERM by enriching the diversity of training domains or better model optimization. Our finding is consistent with the view in work \cite{sener2022domain}. It shows that for numerous penalty-based method recipe that solves a joint optimization problem of minimizing the empirical risk with the surrogate penalty, a significant failure mode is an excess risk in the joint optimization, which fails to optimize the empirical risk (i.e., in-distribution performance) sufficiently compared with ERM.

Under the ``ST (swap test)" evaluation criterion, we have the following findings. 1) On the Chinese character recognition dataset, most of the tested methods are susceptible to the selection of training domains (see \myfigref{PaHCC}). Among them, CORAL and IB-ERM have stable performance advantages. 2) On the object recognition dataset, the relative performance of the tested methods is mostly stable (see \myfigref{DomainNet_0}). Among them, Fish and CORAL have stable performance advantages. 3) In general, CORAL is a more stable domain generalization method compared with other tested methods.
\section{Conclusion}
This paper presents a practical and challenging domain generalization task for the handwritten Chinese character recognition scenario and constructs a large-scale Non-I.I.D. image dataset called PaHCC with 1000 categories and 996478 samples. Our extensive empirical evaluations of DG methods on PaHCC and DomainNet datasets lead to four conclusions. First, Chinese character recognition differs from object recognition on distribution shifts and is a more challenging generalization scenario. Second, the existing DG methods do not solve this task well, and there is still much room for improvement. Third, the common leave-one-out protocol is unreliable, which is mainly manifested in the following two aspects. 1) The selection of training domains affects the performance of DG methods without changing the number of training domains. 2) The performance of some methods deteriorates when increasing the number of training domains. Fourth, the proposed ``dynamic" DG setting reveals more properties of DG methods, which assist the original criterion for more reliable evaluations. Going forward, we hope that our dataset and results could promote comprehensive and reliable evaluation and enable advances in domain generalization.
 

\bibliography{mybibfile}

\begin{thebibliography}{10}
\expandafter\ifx\csname url\endcsname\relax
  \def\url#1{\texttt{#1}}\fi
\expandafter\ifx\csname urlprefix\endcsname\relax\def\urlprefix{URL }\fi
\expandafter\ifx\csname href\endcsname\relax
  \def\href#1#2{#2} \def\path#1{#1}\fi

\bibitem{torralba_unbiased_2011}
A.~Torralba, A.~A. Efros, Unbiased look at dataset bias, in: CVPR, 2011, pp. 1521--1528.

\bibitem{ZhouLQXL23}
K.~Zhou, Z.~Liu, Y.~Qiao, T.~Xiang, C.~C. Loy, Domain generalization: {A} survey, IEEE Transactions on Pattern Analysis and Machine Intelligence 45~(4) (2023) 4396--4415.

\bibitem{wang2022generalizing}
J.~Wang, C.~Lan, C.~Liu, Y.~Ouyang, T.~Qin, W.~Lu, Y.~Chen, W.~Zeng, P.~Yu, Generalizing to unseen domains: A survey on domain generalization, IEEE Transactions on Knowledge and Data Engineering.

\bibitem{GulrajaniL21}
I.~Gulrajani, D.~Lopez{-}Paz, In search of lost domain generalization, in: ICLR, 2021.

\bibitem{KohSMXZBHYPGLDS21}
P.~W. Koh, S.~Sagawa, H.~Marklund, S.~M. Xie, M.~Zhang, A.~Balsubramani, W.~Hu, M.~Yasunaga, R.~L. Phillips, I.~Gao, T.~Lee, E.~David, I.~Stavness, W.~Guo, B.~Earnshaw, I.~S. Haque, S.~M. Beery, J.~Leskovec, A.~Kundaje, E.~Pierson, S.~Levine, C.~Finn, P.~Liang, {WILDS:} {A} benchmark of in-the-wild distribution shifts, in: ICML, Vol. 139, 2021, pp. 5637--5664.

\bibitem{zhang2011style}
X.-Y. Zhang, C.-L. Liu, Style transfer matrix learning for writer adaptation, in: CVPR, 2011, pp. 393--400.

\bibitem{zhang2012writer}
X.-Y. Zhang, C.-L. Liu, Writer adaptation with style transfer mapping, IEEE transactions on pattern analysis and machine intelligence 35~(7) (2012) 1773--1787.

\bibitem{Ian_deep_learning2016}
I.~J. Goodfellow, Y.~Bengio, A.~C. Courville, \href{http://www.deeplearningbook.org/}{Deep Learning}, Adaptive computation and machine learning, {MIT} Press, 2016.
\newline\urlprefix\url{http://www.deeplearningbook.org/}

\bibitem{ShankarPCCJS18}
S.~Shankar, V.~Piratla, S.~Chakrabarti, S.~Chaudhuri, P.~Jyothi, S.~Sarawagi, Generalizing across domains via cross-gradient training, in: ICLR, 2018.

\bibitem{Volpi_generalizing_2018}
R.~Volpi, H.~Namkoong, O.~Sener, J.~C. Duchi, V.~Murino, S.~Savarese, Generalizing to unseen domains via adversarial data augmentation, in: NeurIPS, 2018, pp. 5339--5349.

\bibitem{zhou_deep_2020}
K.~Zhou, Y.~Yang, T.~M. Hospedales, T.~Xiang, Deep domain-adversarial image generation for domain generalisation, in: AAAI, 2020, pp. 13025--13032.

\bibitem{RobeyPH21}
A.~Robey, G.~J. Pappas, H.~Hassani, Model-based domain generalization, in: NeurIPS, 2021, pp. 20210--20229.

\bibitem{LiLLGFH21}
P.~Li, D.~Li, W.~Li, S.~Gong, Y.~Fu, T.~M. Hospedales, A simple feature augmentation for domain generalization, in: ICCV, 2021, pp. 8866--8875.

\bibitem{zhou_domain_2021}
K.~Zhou, Y.~Yang, Y.~Qiao, T.~Xiang, Domain generalization with mixstyle, in: ICLR, 2021.

\bibitem{KangLKK22}
J.~Kang, S.~Lee, N.~Kim, S.~Kwak, Style neophile: Constantly seeking novel styles for domain generalization, in: CVPR, 2022, pp. 7120--7130.

\bibitem{0001GXL21}
J.~Huang, D.~Guan, A.~Xiao, S.~Lu, {FSDR:} frequency space domain randomization for domain generalization, in: CVPR, 2021, pp. 6891--6902.

\bibitem{XuZ0W021}
Q.~Xu, R.~Zhang, Y.~Zhang, Y.~Wang, Q.~Tian, A fourier-based framework for domain generalization, in: CVPR, 2021, pp. 14383--14392.

\bibitem{Muandet_domain_2013}
K.~Muandet, D.~Balduzzi, B.~Sch{\"{o}}lkopf, Domain generalization via invariant feature representation, in: ICML, Vol.~28, 2013, pp. 10--18.

\bibitem{LiPWK18}
H.~Li, S.~J. Pan, S.~Wang, A.~C. Kot, Domain generalization with adversarial feature learning, in: CVPR, 2018, pp. 5400--5409.

\bibitem{LiTGLLZT18}
Y.~Li, X.~Tian, M.~Gong, Y.~Liu, T.~Liu, K.~Zhang, D.~Tao, Deep domain generalization via conditional invariant adversarial networks, in: ECCV, Vol. 11219, 2018, pp. 647--663.

\bibitem{li_conditional_2018}
Y.~Li, M.~Gong, X.~Tian, T.~Liu, D.~Tao, Domain generalization via conditional invariant representations, in: AAAI, 2018, pp. 3579--3587.

\bibitem{peng_domain_2019}
X.~Peng, Z.~Huang, X.~Sun, K.~Saenko, Domain agnostic learning with disentangled representations, in: ICML, Vol.~97, 2019, pp. 5102--5112.

\bibitem{bai_decaug_2021}
H.~Bai, R.~Sun, L.~Hong, F.~Zhou, N.~Ye, H.~Ye, S.~G. Chan, Z.~Li, Decaug: Out-of-distribution generalization via decomposed feature representation and semantic augmentation, in: AAAI, 2021, pp. 6705--6713.

\bibitem{Arjovsky_Invariant_2019}
M.~Arjovsky, L.~Bottou, I.~Gulrajani, D.~Lopez{-}Paz, Invariant risk minimization, arXiv preprint arXiv:1907.02893.

\bibitem{NamLPYY21}
H.~Nam, H.~Lee, J.~Park, W.~Yoon, D.~Yoo, Reducing domain gap by reducing style bias, in: CVPR, 2021, pp. 8690--8699.

\bibitem{li_learning_2018}
D.~Li, Y.~Yang, Y.~Song, T.~M. Hospedales, Learning to generalize: Meta-learning for domain generalization, in: AAAI, 2018, pp. 3490--3497.

\bibitem{ding_deep_2018}
Z.~Ding, Y.~Fu, Deep domain generalization with structured low-rank constraint, {IEEE} Transactions on Image Processing 27~(1) (2018) 304--313.

\bibitem{seo_learning_2020}
S.~Seo, Y.~Suh, D.~Kim, G.~Kim, J.~Han, B.~Han, Learning to optimize domain specific normalization for domain generalization, in: ECCV, Vol. 12367, 2020, pp. 68--83.

\bibitem{GaninUAGLLML16}
Y.~Ganin, E.~Ustinova, H.~Ajakan, P.~Germain, H.~Larochelle, F.~Laviolette, M.~Marchand, V.~S. Lempitsky, Domain-adversarial training of neural networks, J. Mach. Learn. Res. 17 (2016) 59:1--59:35.

\bibitem{KimYPKL21}
D.~Kim, Y.~Yoo, S.~Park, J.~Kim, J.~Lee, Selfreg: Self-supervised contrastive regularization for domain generalization, in: ICCV, 2021, pp. 9599--9608.

\bibitem{huang_selfChallenge_2020}
Z.~Huang, H.~Wang, E.~P. Xing, D.~Huang, Self-challenging improves cross-domain generalization, in: ECCV, Vol. 12347, 2020, pp. 124--140.

\bibitem{sagawa2019distributionally}
S.~Sagawa, P.~W. Koh, T.~B. Hashimoto, P.~Liang, Distributionally robust neural networks, in: ICLR, 2020.

\bibitem{ParascandoloNOG21}
G.~Parascandolo, A.~Neitz, A.~Orvieto, L.~Gresele, B.~Sch{\"{o}}lkopf, Learning explanations that are hard to vary, in: ICLR, 2021.

\bibitem{WangLK20}
Y.~Wang, H.~Li, A.~C. Kot, Heterogeneous domain generalization via domain mixup, in: ICASSP, 2020, pp. 3622--3626.

\bibitem{SunS16}
B.~Sun, K.~Saenko, Deep {CORAL:} correlation alignment for deep domain adaptation, in: ECCV Workshops, Vol. 9915, 2016, pp. 443--450.

\bibitem{KruegerCJ0BZPC21}
D.~Krueger, E.~Caballero, J.~Jacobsen, A.~Zhang, J.~Binas, D.~Zhang, R.~L. Priol, A.~C. Courville, Out-of-distribution generalization via risk extrapolation (rex), in: ICML, Vol. 139, 2021, pp. 5815--5826.

\bibitem{AhujaCZGBMR21}
K.~Ahuja, E.~Caballero, D.~Zhang, J.~Gagnon{-}Audet, Y.~Bengio, I.~Mitliagkas, I.~Rish, Invariance principle meets information bottleneck for out-of-distribution generalization, in: NeurIPS, 2021, pp. 3438--3450.

\bibitem{HuangWXH20}
Z.~Huang, H.~Wang, E.~P. Xing, D.~Huang, Self-challenging improves cross-domain generalization, in: ECCV, Vol. 12347, 2020, pp. 124--140.

\bibitem{shahtalebi2021sand}
S.~Shahtalebi, J.-C. Gagnon-Audet, T.~Laleh, M.~Faramarzi, K.~Ahuja, I.~Rish, Sand-mask: An enhanced gradient masking strategy for the discovery of invariances in domain generalization, arXiv preprint arXiv:2106.02266.

\bibitem{PezeshkiKBCPL21}
M.~Pezeshki, S.~Kaba, Y.~Bengio, A.~C. Courville, D.~Precup, G.~Lajoie, Gradient starvation: {A} learning proclivity in neural networks, in: NeurIPS, 2021, pp. 1256--1272.

\bibitem{ShiSTNHUS22}
Y.~Shi, J.~Seely, P.~H.~S. Torr, S.~Narayanaswamy, A.~Y. Hannun, N.~Usunier, G.~Synnaeve, Gradient matching for domain generalization, in: ICLR, 2022.

\bibitem{li2017deeper}
D.~Li, Y.~Yang, Y.~Song, T.~M. Hospedales, Deeper, broader and artier domain generalization, in: ICCV, 2017, pp. 5543--5551.

\bibitem{fang_unbiased_2013}
C.~Fang, Y.~Xu, D.~N. Rockmore, Unbiased metric learning: {On} the utilization of multiple datasets and web images for softening bias, in: Proceedings of the {IEEE} {International} {Conference} on {Computer} {Vision}, 2013, pp. 1657--1664.

\bibitem{PengBXHSW19}
X.~Peng, Q.~Bai, X.~Xia, Z.~Huang, K.~Saenko, B.~Wang, Moment matching for multi-source domain adaptation, in: ICCV, 2019, pp. 1406--1415.

\bibitem{he_towards_2021}
Y.~He, Z.~Shen, P.~Cui, Towards non-iid image classification: A dataset and baselines, Pattern Recognition 110 (2021) 107383.

\bibitem{venkateswara2017deep}
H.~Venkateswara, J.~Eusebio, S.~Chakraborty, S.~Panchanathan, Deep hashing network for unsupervised domain adaptation, in: CVPR, 2017, pp. 5385--5394.

\bibitem{lecun1998gradient}
Y.~LeCun, L.~Bottou, Y.~Bengio, P.~Haffner, Gradient-based learning applied to document recognition, Proceedings of the IEEE 86~(11) (1998) 2278--2324.

\bibitem{LiuYWW11}
C.~Liu, F.~Yin, D.~Wang, Q.~Wang, {CASIA} online and offline chinese handwriting databases, in: ICDAR, 2011, pp. 37--41.

\bibitem{LiuYWW13}
C.~Liu, F.~Yin, D.~Wang, Q.~Wang, Online and offline handwritten chinese character recognition: Benchmarking on new databases, Pattern Recognition 46~(1) (2013) 155--162.

\bibitem{ZhouLZZ22}
X.~Zhou, Y.~Lin, W.~Zhang, T.~Zhang, Sparse invariant risk minimization, in: ICML, Vol. 162, 2022, pp. 27222--27244.

\bibitem{sener2022domain}
O.~Sener, V.~Koltun, Domain generalization without excess empirical risk, NeurIPS 35 (2022) 13380--13391.

\end{thebibliography}

\end{document}